\documentclass[11pt]{article}

\usepackage[preprint]{acl}
\usepackage{times}
\usepackage{latexsym}
\usepackage[utf8]{inputenc}
\usepackage{pifont}
\usepackage{hyperref}
\usepackage{booktabs} 
\usepackage{amsmath}
\usepackage{tcolorbox}
\usepackage{microtype}
\usepackage{graphicx}

\usepackage{array}
\usepackage{xcolor}
\usepackage{colortbl}
\usepackage{siunitx}
\usepackage{pdflscape}   

\definecolor{headerblue}{RGB}{30,70,130}
\definecolor{rowgray}{RGB}{245,247,250}

\begin{document}

\newtcolorbox{mybox}[1]{
  colback=cyan!5!white,
  colframe=cyan!75!black,
  fonttitle=\bfseries,
  title=#1,
  left=3pt,
  right=3pt,
  top=2pt,
  bottom=2pt
}

\title{GPTKB Studio: Building Disambiguated Knowledge Bases\\from Any Language Model, on Any Topic, On the Fly}

\author{
  Luca Giordano \quad \quad Yujia Hu \quad \quad Simon Razniewski \\
  ScaDS.AI Dresden/Leipzig \& TU Dresden, Germany \\
  \texttt{\{luca.giordano,yujia.hu,simon.razniewski\}@tu-dresden.de}
}

\maketitle

\begin{abstract}
Knowledge materialization shifts the study of LLM knowledge from answering predefined benchmark questions to systematically extracting entire knowledge bases (KBs). Yet, existing approaches have largely confined materialization to offline processes, making KBs available only as precomputed artifacts. We present GPTKB Studio, an open-source, interactive web system for knowledge materialization. Users can configure materializations in their browser, specifying an OpenAI-compatible endpoint, topic, seed entity, language, prompts, stopping criteria, and separate models for each phase. They can monitor progress in real time, explore the resulting KB as interlinked HTML pages, or download it as a SQL database. The backend inherits the on-the-fly disambiguation pipeline of GPTKB~2.0 \citep{hu2026gptkb20}, producing canonicalized entities, relations, and classes with fully auditable disambiguation decisions. We evaluate speed, scalability, and domain-specific disambiguation accuracy, demonstrating that GPTKB Studio can rapidly construct large, accurately disambiguated, domain-specific KBs.\\
 {\scriptsize System: \href{https://gptkbstudio.scads.ai/}{gptkbstudio.scads.ai}}\\[-2pt]
{\scriptsize Code:  \href{https://github.com/Knowledge-aware-AI/GPTKB-Studio}{github.com/Knowledge-aware-AI/GPTKB-Studio}}\\[-2pt]
{\scriptsize Video: \href{https://youtu.be/p3mAnqfOZUg}{youtu.be/p3mAnqfOZUg}}
\end{abstract}

\section{Introduction}
\begin{figure}[h!]
    \centering
    \includegraphics[width=1\linewidth]{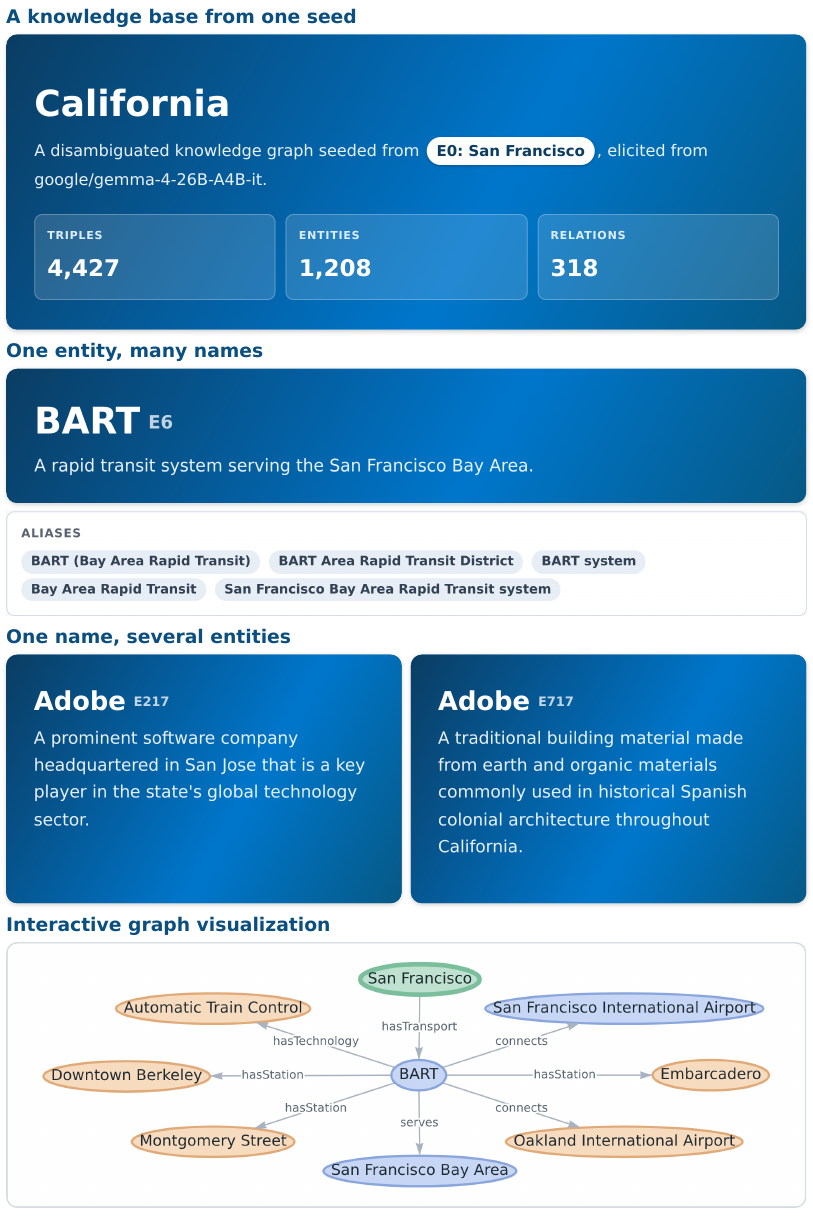}
    \caption{Excerpts from the report of a materialization on \emph{California} seeded with \emph{San Francisco} (Gemma~4): five names the model used for BART were merged into one entity; two entities named \emph{Adobe} (the software company and the building material) were kept apart.}
    \label{fig:teaser}
\end{figure}

What Large Language Models (LLMs) know is measured almost exclusively through benchmarks: fixed sets of questions whose answers are known in advance. Such benchmarks inherit an
\emph{availability bias} \citep{tversky1973availability}: they can only reveal
knowledge that somebody thought to ask about
\citep{bean2025measuring,hu2025enabling}. A model may hold a million facts
about topics no benchmark designer ever considered, and no amount of
benchmarking will surface them.

\emph{Knowledge materialization} takes the opposite approach. Starting from a
single seed entity, the model is asked to state everything it knows about that
entity as subject-predicate-object triples; named entities appearing among the
objects become new subjects; the process recurses
\citep{cohen2023crawling,hu2025enabling}. The most recent implementations of this approach include real-time entity and relation disambiguation at inference time \citep{hu2026gptkb20}. What comes out is a disambiguated knowledge base of the model's beliefs that can be browsed, queried,
and analyzed with ordinary database technology. 
\citet{razniewski2026epistemic} argue that such \emph{epistemic twins} should
eventually ship alongside models as a standard auditing artifact, much like model
cards \citep{mitchell2019model}.

However, so far the research community met this paradigm only through public code repositories and its outputs \citep{hu2026gptkb15,hu2026gptkb20demo}. Three large
materializations cost \$3.5k, more than \$14k and \$7k respectively \citep{hu2025enabling,hu2026gptkb15,hu2026gptkb20},
and each was executed once, with one model, from one seed, in one language.

This work presents GPTKB Studio, an open-source, interactive tool for on-the-fly knowledge materialization from LLMs. Its contributions are:

\begin{enumerate}\itemsep0pt
    \item \textbf{From artifact to factory.} Existing demos in this line let users browse a knowledge base that somebody else built \cite{hu2026gptkb15,hu2026gptkb20demo}. GPTKB Studio lets \textit{anyone} build one, for \textit{any topic}, from \textit{any OpenAI-compatible endpoint}, from any model, in any language.
    \item \textbf{An observable, steerable materialization.} GPTKB Studio streams metrics and logs as the materialization runs, and allows interactively pausing, resuming, stopping, with hard budget caps on entities, depth, runtime and cost. Phenomena previously reported as experimental observations (e.g. non-termination, synonym explosion, degenerate entity loops; \citealp{giordano2026foundations}) become directly visible.
    \item \textbf{Disambiguated by construction, and auditable end to end.} The backend inherits the context-guided disambiguation pipeline of GPTKB~2.0 \citep{hu2026gptkb20}, with separate, user-selectable models for the three phases of elicitation, named entity recognition and disambiguation, and a user-controlled trade-off between merging and splitting. Every merge decision it makes, with the candidates and context the model saw, can be inspected.
    \item \textbf{Results that are immediately usable.} Each materialization produces an interlinked HTML collection (statistics, an interactive graph, one page per entity) that can be opened and shared as-is, plus an SQL database for downstream use.
\end{enumerate}

\noindent We describe the system and its implementation (\S\ref{fetures_implementation}); the demonstration (\S\ref{use_cases}); the evaluation (\S\ref{sec:eval}); and related work (\S\ref{rw}).

\section{System Overview}
\label{fetures_implementation}

\begin{figure*}[t]
\centering
\includegraphics[width=\textwidth]{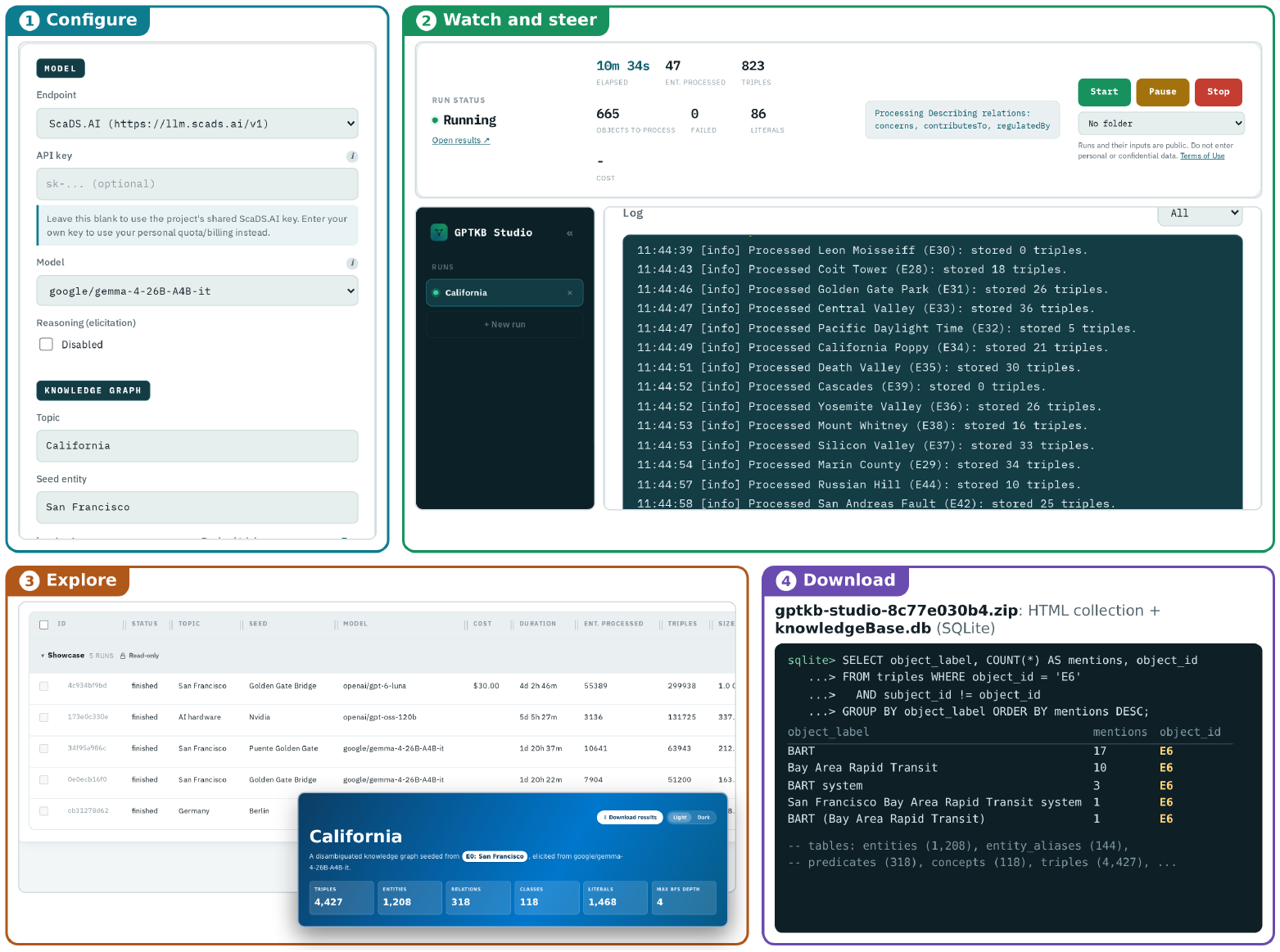}
\caption{The four steps of a GPTKB Studio session, on a materialization of \emph{California} seeded with \emph{San Francisco} (for illustrative purposes only, capped before organic termination). (1)~\emph{Configure}: endpoint, API key, model, topic, seed entity, language and the number of facts requested per entity. (2)~\emph{Watch and steer}: live metrics (elapsed time, explored entities, triples, objects still to process, failures, literals, cost), the current pipeline step, controls to pause and stop, and the streaming log. (3)~\emph{Explore}: the run explorer lists all materializations in folders, including a read-only showcase; each materialization opens as an HTML report (here, the California materialization's overview page). (4)~\emph{Download}: the archive contains the HTML collection and a SQL database.}
\label{fig:overview}
\end{figure*}

\subsection{Configuring a materialization}

A GPTKB Studio session consists of four steps. The user (i) \emph{configures} a materialization in a
single form (Part 1 in Figure~\ref{fig:overview}), (ii) \emph{watches and steers} it
while metrics and a log update live (Part 2 in Figure~\ref{fig:overview}), (iii) \emph{explores} the result in a generated
HTML collection (Part 3 in Figure~\ref{fig:overview}), and (iv) \emph{downloads} it as database (Part 4 in Figure~\ref{fig:overview}). All materializations are archived and inspectable in a run explorer.

\paragraph{Models.}
Any OpenAI-compatible endpoint can be used: presets cover OpenAI, OpenRouter
and our local ScaDS.AI endpoint, and a custom URL allows for
self-hosted deployments (e.g.\ vLLM). Following GPTKB~2.0, the user picks one model, or, under the advanced options, a different model for each of three tasks: for \emph{elicitation}, for \emph{named entity recognition} (NER), and
for \emph{named entity disambiguation} (NED), the latter also covering
relations, classes and description generation. NER and NED models can come
from a different endpoint than the elicitation model, so that, e.g., a commercial
model's knowledge can be materialized while the many short NER and NED calls
run on local open-weight models.

\paragraph{Knowledge graph, prompt and budget.}
The \emph{topic} confines the materialization to a specific region of the model's
knowledge; the \emph{seed entity} is where it starts (see Part 1 in Figure~\ref{fig:overview}). Eleven prompt languages
are offered pre-packaged, with the prompt already translated; the user can still choose to write the elicitation prompt themselves in any language desired. A three-step slider (few, medium, many) scales the facts requested per entity.
The assembled prompt is editable and stored with the run, and three schema
modes are available: \emph{open}, \emph{example-driven} and \emph{fixed
ontology}. Four caps (entities, BFS depth, runtime, cost) can be combined;
entity and depth caps stop the exploration but let the materialization finish
disambiguating what it has elicited, so a capped run still yields a consistent
knowledge base.
Additional advanced options include the number of parallel requests, maximum requests per minute, and the number of candidates shown to the model during NED.

\subsection{Disambiguation}
\label{sec:disamb}

Materializations that identify entities solely by their surface strings conflate homonyms
(\emph{Paris} the city and \emph{Paris} the Trojan prince) and fragment synonyms (\emph{Marduk Temple}, \emph{Temple of
Marduk}); \citet{giordano2026foundations} found the latter to underlie low
lexical reproducibility and, in some languages, non-termination. GPTKB Studio
therefore implements the pipeline of GPTKB~2.0 \citep{hu2026gptkb20}, in which
every entity, relation and class has an identifier, a label, a description and
aliases\footnote{\href{https://github.com/Knowledge-aware-AI/GPTKB2_disambiguation_on_the_fly}{GPTKB 2.0 GitHub repository.}}.

\paragraph{Pipeline.}
The pipeline proceeds in rounds, each running every step on the items ready for
it. \emph{Elicitation} gives the model the entity's description along with its
label, so that homonyms are elicited as the entity they denote. \emph{NER}
classifies each triple's object as entity or literal. \emph{Triple-context NED}
shows the model the object, its triple, and the most similar existing entities
(by label embedding) with their descriptions; the model matches one, declares
the object new, or abstains. New objects get a generated description and a
second, \emph{description-context} NED pass. Relations and classes go through
the same steps. Caching and guarded parallelization (similar pending labels
are disambiguated one per round, so that two mentions of a new entity cannot
race into two entities) follow GPTKB~2.0 as well. GPTKB Studio adapts the pipeline
to topic-specific materializations by only exploring entities relevant to the topic and by generating descriptions with the topic as
context (Appendix~\ref{app:deviations}).

\paragraph{Controlling merging.}
Merging and splitting are the two directions in which disambiguation can err
\citep{hu2026gptkb20}. GPTKB Studio implements a new \emph{entity
merging} slider in its advanced options (Appendix~\ref{app:deviations}): at the standard setting, the NED model sees the most similar
existing entities and decides; towards \emph{cautious}, an entity is offered as
a match only if it has at least a given similarity to the mention (e.g.\
90\%), so fewer but safer merges are made, at the cost of more duplicates.

\subsection{Running on shared, rate-limited endpoints}
\label{sec:limits}

Disambiguation multiplies the number of model calls, and endpoints might limit requests. As a new feature for the pipeline (Appendix~\ref{app:deviations}), GPTKB Studio's backend reads each model's limits from the
endpoint's rate-limit headers and paces its requests just below them.

\subsection{Watching and steering a materialization}

The dashboard (Part 2 in Figure \ref{fig:overview}) shows elapsed time, explored entities, triples, objects to process, literals, failures and cost, the current step and a log. Materializations can be
paused, stopped and resumed, possibly with other models or a larger budget.
Termination \citep{giordano2026foundations} can be observed live:
a healthy materialization's queue rises and drains; a degenerate one keeps growing indefinitely, with
a log full of near-duplicate names.

\subsection{Output}

Every materialization yields a self-contained \textbf{HTML collection}: statistics, configuration and prompt, a
disambiguation section, an interactive graph visualization, distributions, and one page per
entity with its triples, aliases, namesakes and disambiguation trace
(every mention resolved to it, with the candidates the model saw), plus a
\textbf{SQLite database} for analyses such as those of \citet{ghosh2025mining}.

\subsection{Architecture}
GPTKB Studio is a Python (FastAPI) backend serving a dependency-light browser
application.\footnote{Plain JavaScript, plus vis-network for graph rendering; no build step.}
Each materialization runs in its own worker process and streams progress to the browser
as server-sent events.

The state of a materialization lives in a per-run SQLite database
that is both its working memory and its export: entities, relations and
classes with identifiers, labels, descriptions and aliases; triples; every
disambiguation decision with its candidates; and failures. Because every step
commits to this database, a materialization survives a server restart, can be resumed
where it stopped, and can be downloaded at any time.

Candidate retrieval for entity disambiguation (see Section \ref{sec:disamb}) uses
label embeddings (Qwen3-Embedding-4B, as in GPTKB~2.0) from an inference API hosted on local hardware; the server needs no GPU or database server, and runs as a hosted instance as well as on a laptop.

API keys entered by users are only kept in memory for the duration of a materialization and used only for requests to the endpoint they were entered for; they are never written to disk, logged, or included in exports. All traffic is encrypted with HTTPS. We recommend using a dedicated key with a spending limit.

\begin{table*}[t]
\centering\footnotesize
\setlength{\tabcolsep}{2.8pt}
\begin{tabular}{lrrrrr|rrrrrr}
\toprule
& \multicolumn{5}{c|}{\textbf{Baseline: without disambiguation}} & \multicolumn{6}{c}{\textbf{GPTKB Studio}} \\
& \textbf{Triples} & \textbf{Entities} & \textbf{T/E} & \textbf{Calls/T} & \textbf{Time} & \textbf{Triples} & \textbf{Labels} & \textbf{Entities} & \textbf{T/E} & \textbf{Calls/T} & \textbf{Time} \\
\midrule
Babylon en$^\dagger$ & 9,533$\pm$1,379 & 1,561$\pm$220 & 6.1 & 0.27 & 42m & 5,191$\pm$197 & 1,195$\pm$39 & 662$\pm$22 & 7.8 & 2.32 & 4h27 \\
Babylon de & 5,382 & 860 & 6.3 & 0.29 & 30m & 3,054 & 756 & 428 & 7.1 & 2.47 & 2h54 \\
Babylon it & 7,263 & 1,155 & 6.3 & 0.28 & 38m & 2,146 & 467 & 284 & 7.6 & 2.46 & 1h56 \\
Babylon en, gpt-oss & 53,136$^*$ & 7,129$^{*\ddagger}$ & 24.4 & 0.07 & 6h00$^*$ & 18,669$^*$ & 745$^*$ & 568$^*$ & 32.9 & 0.58 & 6h00$^*$ \\
TBBT en$^\dagger$ & 351$\pm$21 & 56$\pm$6 & 6.3 & 0.28 & 1m & 1,256$\pm$222 & 207$\pm$34 & 171$\pm$34 & 7.4 & 2.42 & 53m \\
DAX~40 en & 3,154 & 633 & 5.0 & 0.32 & 12m & 481 & 79 & 62 & 7.8 & 2.31 & 18m \\
Radebeul en & 21 & 6 & 3.5 & 0.48 & 11s & 41 & 15 & 14 & 2.9 & 3.46 & 2m \\
Babylon en, cautious & & & & & & 8,723 & 1,729 & 1,104 & 7.9 & 2.04 & -- \\
\bottomrule
\end{tabular}
\caption{Evaluation results (elicitation: Llama-3.3-70B unless noted; NER: gemma-4-26B; NED: Qwen3.8-27B).
Labels: names merged into the entities; T/E: triples per explored entity; Calls/T: model calls per triple.
$^\dagger$Mean $\pm$ sd over three runs. $^*$Stopped by the 6-hour cap. $^\ddagger$2,175 explored.}
\label{tab:runs}
\end{table*}

\section{Demonstration Experience}
\label{use_cases}

The conference demonstration consists of three parts. \href{https://youtu.be/p3mAnqfOZUg}{Here} is a screen recording showcasing and describing the system in use.

\paragraph{A. Building a custom KB.} Visitors will use GPTKB Studio to generate KBs on the spot, e.g. about San Francisco or NAACL. Leaving advanced settings as default, together with the visitor we will choose a model, a topic, a seed, and start the materialization. We will then explain the streaming log that shows how the KB grows in real time. The visitors will then explore the result via the generated HTML report, looking at the main information e.g. count of entities and facts, the graph visualization, and top classes and relations. Then, we will navigate the actual triples, focusing on how the recursion progressed from the seed entity (e.g., ``Golden Gate Bridge'' to ``Alcatraz'' to ``Al Capone'').
\paragraph{B. Explore and compare pre-built KBs.} Visitors will find high-quality, pre-built, curated KBs in GPTKB Studio's Run Explorer: a KB on \emph{Germany} built by Gemma 4 26B, a KB on \emph{AI Hardware} built by GPT-OSS-120B, and three KBs on \emph{San Francisco}, two of which built by Gemma 4 26B in English and Spanish respectively, the third built by GPT-6-Luna with a budget of 30\$. These KBs will engage visitors on a treasure hunt looking for e.g. the most verbose model (``Does Gemma 4 26B know more facts about San Francisco than GPT-6-Luna?"), errors (e.g. GPT-6-Luna believes that the Transamerica Pyramid, a renowned building in San Francisco, has two wings near its base instead of at its top\footnote{\texttt{(Transamerica Pyramid, hasArchitecturalFeature, "Two wings near the base")}}), or differences across languages within the same topic and model combination (e.g. Gemma 4 26B generated 63.9K facts about San Francisco in Spanish while only 51.2K in English). Furthermore, visitors will engage in a deeper analysis of the results report and the generated KBs, e.g. by investigating the fully auditable disambiguation decisions, observing the ones that the model got right or wrong, and judging themselves on ambiguous ones (e.g. does \emph{Soraya Taheri}, a character in Khaled Hosseini's \emph{The Kite Runner}, appear in the 2003 novel or in its 2007 film adaptation? GPT-6-Luna could not decide between the two candidate entities).
\paragraph{C. Diving in-depth into advanced configuration settings.} In the third part, visitors will dive into GPTKB Studio's advanced configuration settings, changing the models for the three different pipeline phases (elicitation, NER, NED), the reasoning switch or effort for reasoning models, the amount of desired triples per entity, different pre-packaged elicitation prompt templates (ranging from precision-oriented, schema-constrained prompts to recall-oriented, open-ended ones), parameters related to performance such as number of parallel requests and maximum requests per minute, or parameters related to disambiguation, such as the amount of candidates shown to the model during NED or the slider that tweaks NED conservativeness or aggressiveness.

\section{Evaluation}
\label{sec:eval}

Earlier work showed that topic-restricted materializations terminate, are stable in
yield and robust to the seed \citep{giordano2026foundations}, and evaluated the
accuracy of what such elicitation produces
\citep{giordano2026foundations,giordano2026bequ}; the disambiguation pipeline
was evaluated at scale \citep{hu2026gptkb20}. Here we
evaluate GPTKB Studio's speed, scalability and domain-specific disambiguation accuracy.

\paragraph{Setup.}
We materialize the three topics analyzed in \cite{giordano2026foundations},  \emph{Babylon},
\emph{The Big Bang Theory}, \emph{DAX~40}, plus a long-tail one
(\emph{Radebeul}), \emph{Babylon} also in German and Italian, with
GPT-OSS-120b, and once with cautious merging, on our local LLM endpoint, with a 6-hour cap (see Table~\ref{tab:runs}). Each
configuration ran both with disambiguation and without it, i.e.\ with the
string-based pipeline of \citet{giordano2026foundations}; \emph{Babylon} and
\emph{The Big Bang Theory} ran three times each. For accuracy, one author
annotated a sample of 541 items, presented blindly (decision, pipeline and
model hidden): 184 NED decisions, stratified by type (merge under another or
the same name, new entity, relation, class); 140 NER decisions; and 217 pairs
of similarly named entities (by label-embedding similarity) or equally named
ones. Brackets give 95\% confidence intervals.

\paragraph{Speed.}
GPTKB Studio itself is not the bottleneck: a run waits for model answers, so
its speed is set by the endpoint's rate limits and the model's latency. On our locally hosted ScaDS.AI endpoint, both pipelines issued model calls at a similar rate (e.g.\ 45
vs.\ 61 per minute for \emph{Babylon}); disambiguation is slower mainly because it
needs about eight times more calls per triple (Table~\ref{tab:runs}). Within
these limits, results appear quickly: the first triples typically arrive within
half a minute of the start, and the first disambiguation decisions came within
6-11\,s in all runs of Table~\ref{tab:runs}; after ten minutes, the
\emph{Germany} materialization (described below) had 420 triples about 16 entities. Pacing
makes the most of a rate-limited endpoint: on a materialization of \emph{The Big Bang
Theory} capped at 60 entities, it avoided all 58 rate-limit errors that
occurred without it, and finished 11\% sooner.

\paragraph{Scalability.}
Our largest materialization, \emph{Germany} seeded with \emph{Berlin} (Gemma-4-26B for
all three tasks), terminated after 6.3 days, with 29,423 entities and
223,237 triples, all disambiguated (7,907 aliases, 3,217 relations, 2,482
classes). Its throughput increased with KB size
(about 1,070 triples per hour in the first 20 hours, and 2,430 in
the last 20). Each run is a
separate process that keeps the knowledge base in SQLite on disk (408\,MB for
\emph{Germany}) and only its vector index in memory: on a virtual machine with
8 vCPUs and 8\,GB of RAM, four materializations ran side by side with 140-210\,MB of
memory each (under 1\,GB including the server) and the CPU almost idle.

\paragraph{Domain-specific disambiguation accuracy.}
KB sizes in Table~\ref{tab:runs} are not directly comparable: facts per
explored entity are preserved (T/E), but GPTKB Studio does not re-explore
merged names.
Disambiguation consolidates 1.1-1.8 surface labels into each entity (e.g.\
1,195 into 662 for \emph{Babylon}), where the pipeline without it keeps variants
apart (\emph{Allianz}, \emph{Allianz SE}, \emph{Allianz Group}). Pooled over all
configurations (strata weighted by size), 78\% [66-89] of merges into an
existing entity were correct ($n$=61), and 82\% [69-95] of entities declared
new had no match among the candidates ($n$=55). Errors mostly merge related but
distinct things (\emph{Sonnengott}, `sun god', into \emph{Shamash}). Among the
most similar entity names (cosine $\geq$\,0.8), none of 69 sampled pairs were
duplicates with disambiguation, against 4 of 68 without, and all 5 pairs of
equally named entities were correctly kept apart. NER was correct for 85\%
[75-95] of objects ($n$=68), against 70\% [50-90] ($n$=70) for the entity
classification of the original pipeline. The cautious merging setting made
merges more precise (87\% vs.\ 73\%) at the cost of more duplicates (1,104 vs.\
662 entities for \emph{Babylon}).

\section{Related Work}
\label{rw}
\citet{cohen2023crawling} first presented the idea of materializing an LLM's internal knowledge
graph; \citet{hu2025enabling} developed this idea into the scalable GPTKB methodology, introducing entity-centric expansion through LLM-based named entity recognition, and reducing the number of prompts per entity. GPTKB~v1.5 \citep{hu2026gptkb15}
repeated the process with the GPT-4.1 frontier LLM and served the result through a
browsing and SPARQL interface. GPTKB~2.0 \citep{hu2026gptkb20} replaced surface
strings with real entity identifiers, resolving synonymy and homonymy on the fly
during construction. \citet{ghosh2025mining} used the resulting
100M-belief resource to analyze GPT-4.1's topical coverage, gender and geographic
biases, hallucination modes, symmetry violations and knowledge cutoff.
Related extraction schemes include list-based polling \citep{hao-etal-2023-bertnet},
schema-guided domain graphs \citep{parovic-etal-2025-generating}, ontology induction
\citep{funk2023towards}, and text-shaped materialization
\citep{saeed2026llmpedia}.

\paragraph{Domain-restricted materializations.}
\citet{giordano2026foundations} introduced miniGPTKBs, domain-restricted
materializations, and answered the methodological questions that the large
materializations had left open: materializations do terminate (for most models and languages), yield is
highly reproducible while surface forms are not, and robustness is high against
seed and temperature but low against language and model choice.

\paragraph{Positioning.}
LLM-based knowledge graph construction tools typically extract knowledge
\emph{from text} \citep{zhang-soh-2024-extract,bai2026autoschemakg}; GPTKB Studio extracts it
\emph{from the model itself}. Its disambiguation is a form of entity matching
\citep{kopcke2010frameworks}, which LLMs now perform directly
\citep{peeters2025entity}; here it runs during construction rather than as a
separate step. Unlike the GPTKB demos
\citep{hu2026gptkb15,hu2026gptkb20demo}, which expose one pre-built knowledge
base, GPTKB Studio exposes the construction process, parameterized and
repeatable.

\section{Conclusion}

Knowledge materialization shows that a language model's factual beliefs can be extracted, inspected and queried. GPTKB Studio makes this possible for anyone: a disambiguated, browsable and auditable KB from any model, on any topic, in any language. We hope GPTKB Studio helps turn epistemic twinning from a demonstration of principle into everyday practice.

\section*{Limitations}

GPTKB Studio inherits the limitations of the methodology it implements.
Materialized knowledge is a \emph{lower bound} on what a model holds: prompt
formulation can affect what is surfaced \citep{petroni2019language,hu2025enabling}, and the definition of LLM knowledge remains a debated topic \cite{fierro2024defining,herrmann2025standards}.

We also emphasize that triple correctness is not guaranteed in the tool: The tool is intended as an inspection tool for LLM beliefs, on purpose including true as well as false beliefs. Validating correctness requires additional external evidence like used in VeriFactScore \cite{liu2025verifact} or BeQu \cite{giordano2026bequ}.





\bibliography{refs.bib}

\appendix

\section{Adaptations of the GPTKB~2.0 pipeline}
\label{app:deviations}

GPTKB Studio keeps the prompts and stage order of GPTKB~2.0 and adapts it to
topic-specific materializations in the following ways.
\begin{itemize}\itemsep0pt
\item \textbf{Elicitation prompt:} the topic-scoped, multilingual prompt of
\citet{giordano2026foundations} is used, so an off-topic subject returns no facts.
\item \textbf{Topic context:} descriptions are generated with the materialization's topic as
context, since a single triple is often not enough to tell which namesake is
meant, and wrong descriptions derail description-context NED.
\item \textbf{Merging control:} an optional similarity floor for candidates
(\S\ref{sec:disamb}); entities with exactly the mention's name are always
offered; the number of candidates (five in GPTKB~2.0) is configurable.
\item \textbf{Pacing:} requests are paced per model below the endpoint's rate
limits (\S\ref{sec:limits}).
\end{itemize}

\section{GPTKB Studio across different models}
\label{app:different_models}

\begin{table}[h]
\centering\small
\setlength{\tabcolsep}{2pt}
\begin{tabular}{lrrrrr}
\toprule
\textbf{Elicitation model} & \textbf{Expl.} & \textbf{T/E} & \textbf{Ent.} & \textbf{Failed} & \textbf{Time} \\
\midrule
Llama-3.3-70B & 14 & 10.4 & 59 & 0 & 11m \\
gemma-4-26B & 17 & 16.3 & 93 & 0 & 20m \\
gpt-oss-120b & 12 & 52.9 & 179 & 0 & 40m \\
Qwen3.8-27B & 14 & 46.4 & 234 & 3 & 59m \\
GLM-5.3 & 26 & 17.3 & 49 & 0 & 1h$^*$ \\
GLM-5.3-Flash & 16 & 57.4 & 104 & 0 & 1h$^*$ \\
MiniMax-M3 & 17 & 78.1 & 106 & 0 & 1h$^*$ \\
DeepSeek-V4.1-Flash & 13 & 106.4 & 98 & 0 & 1h$^*$ \\
\bottomrule
\end{tabular}
\caption{Materializations of \emph{the ancient city of Babylon} capped at 30 entities
with different elicitation models (NER: gemma-4-26B, NED: Qwen3.8-27B).
Expl.: explored entities; T/E: triples per explored entity; Ent.: entities;
Failed: entities given up after repeated failures (Qwen3.8: empty or
repetitive answers). $^*$Stopped by the 1-hour cap with facts still awaiting
disambiguation (15 requests per minute on the endpoint).}
\label{tab:different_models}
\end{table}

\end{document}